\documentclass{article}
\usepackage{spconf,amsmath,epsfig}

\newcommand{\ignore}[1]{}

\begin{document}\sloppy

\def\x{{\mathbf x}}
\def\L{{\cal L}}

\title{SnowWatch: Snow Monitoring through Acquisition and Analysis of User-Generated Content}
%
\name{Roman Fedorov, Piero Fraternali, Chiara Pasini, Marco Tagliasacchi}
\address{Politecnico di Milano, Dipartimento di Elettronica, Informazione e Bioingegneria\\Piazza Leonardo da Vinci, 32 - 20133 Milan - Italy}

\maketitle

\let\thefootnote\relax\footnotetext{Author Emails: \{first\}.\{last\}@polimi.it\\This work is supported by the EU FP7 CUbRIK Integrating Project, http://www.cubrikproject.eu and POR-FESR 2007-2013 PROACTIVE Project, http://www.proactiveproject.eu}

\begin{abstract}
We present a system for complementing snow phenomena monitoring with virtual measurements extracted from public visual content. The proposed system integrates an automatic acquisition and analysis of photographs and webcam images depicting Alpine mountains. In particular, the technical demonstration consists in a web portal that interfaces the whole system with the population. It acts as an entertaining photo-sharing social web site, acquiring at the same time visual content necessary for environmental monitoring.
\end{abstract}
\begin{keywords}
UGC, photo-sharing, environmental sensing, snow monitoring
\end{keywords}
\section{Introduction}

In a period of climate change and shrinking public investments in monitoring infrastructures, the need of a low cost analysis of environmental and ecological parameters is extremely important. Several state-of-the-art methods that try to virtualize the permanent measurement stations using mountain image analysis~\cite{garvelmann2013observation}, using both ground and aerial images were recently proposed. These methods suffer from the absence or high cost of input data and are insufficient to produce and calibrate a really usable mountain environmental model. On the other hand, the amount of visual User-Generated Content publicly available on the Web is reaching an unprecedented size. A significant portion of this data consists in public photographs that often depict outdoor scenes. We argue that publicly available visual content, in the form of user-generated photographs and image feeds from outdoor webcams, can both be leveraged as additional measurement sources, complementing existing ground, satellite and airborne sensor data. In particular we address the problem of snow monitoring in the Alpine regions.

The proposed technical demonstration is an extension of~\cite{fedorov2014snow}, where several independent components were presented from technical perspective. This work consists in an web-accessible integrated system which acts as a central collection point for public media content depicting Alpine mountains, and its analysis aimed at the extraction of snow cover data.
Another challenge of the project is boosting social engagement for environmental monitoring. 

\section{Scientific and Technical Description}

The first goal of the system is to acquire a large amount of relevant media content. This objective is reached by means of tree different processes:
\textbf{1)~Relevant photograph retrieval} from photo-sharing platforms and social networks. Given the source of the photographs all the images available in the defined area along with their geo-tags are retrieved. A filtering based on the altitude of the photographer during the shot is then applied. A supervised learning Support Vector Machine performs then a two-class content-based classification, discarding the photos that do not contain clear mountain profiles.
\textbf{2)~Relevant webcam images retrieval} through constant feed acquisition. A webcam placed in a mountain region is considered as a source of relevant photos with a high temporal resolution. Cloudy meteorological conditions, however, are very common to be found at high altitudes, therefore only images with good weather conditions are automatically identified and aggregated daily. 
\textbf{3)~Relevant photograph uploading} performed by the users engaged directly in the environmental monitoring using the photo-sharing web platform.


Estimating snow cover index from the media content consists in two steps:
\textbf{1)~Photograph mountain identification}: given as input a geo-tagged photograph, estimate the direction of the camera using a matching algorithm on the photograph edge maps and
a rendered view of the mountain silhouettes that should be seen from the observer’s point of view~\cite{fedorov2014mountain}. Once the direction of the photograph is known the snow coverage estimation component is invoked.
\textbf{2)~Snow cover area identification}: the portion of the photograph representing the mountain can be extracted by considering only the pixels of the rendered Digital Elevation Model view that correspond to the terrain points at an altitude above a defined threshold. Each pixel of the image representing terrain is then automatically label as covered by snow or not.

\section{Implementation and Use}

Beside the acquisition of the dataset for the environmental analysis, the proposed technical demonstration appears to the user as an entertainment social photo-sharing platform. The platform provides tools for easy retrieval of multi-source mountain media and sharing of own photographs with other mountaineers.

Given an alignment with the rendered terrain view and an estimated snow/no-snow mask, several information can be obtained for each pixel of the photograph, among which whether the pixel represents sky or terrain, what is the altitude of the point above the sea level, how distant was the point from the photographer, and which points corresponds to mountain peaks.
The idea is to exploit this information to enrich the user experience of exploring mountain photographs, allowing the user to switch from simple view of a photo to a interactive experience based on the 3D model of the terrain. Another important point of social engagement lies in the fact that the alignment that brings to the photograph orientation estimation and peak identification can be wrong. This can occur due to slightly incorrect photograph geo-tag or to challenging conditions such as bad weather or partial skyline occlusion. The user so can view the alignment between the photograph and the terrain rendered view, and eventually correct it. Once a new alignment is provided by the user, the system recomputes the snow estimation based on the new photograph direction of view.

\begin{figure}[h!]
  \centering
    \includegraphics[width=1.0\linewidth]{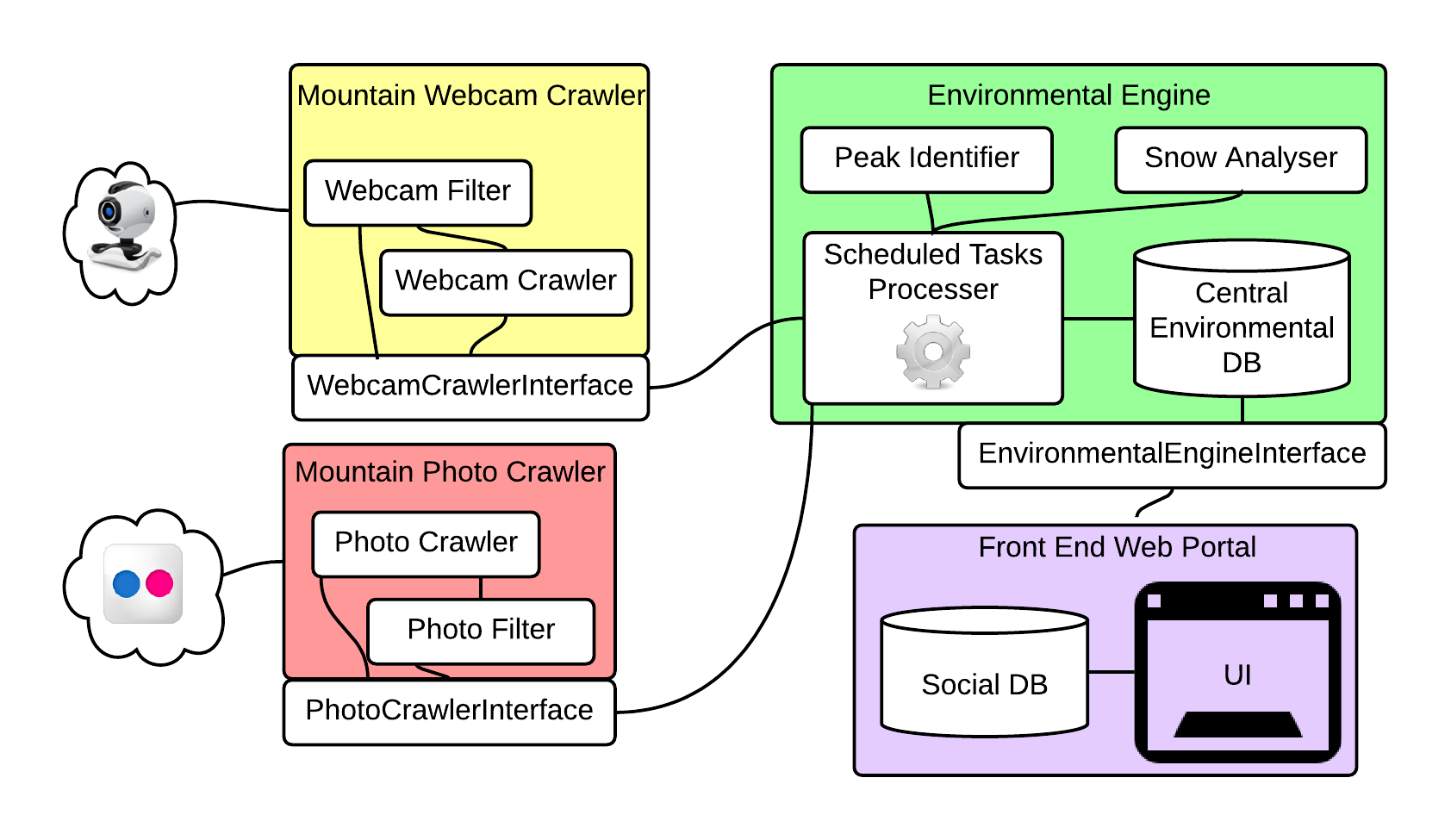}
  \caption{Architecture schema of the system.}
  \label{figure:schema}    
\end{figure}

Figure~\ref{figure:schema} shows the schematic architecture overview of the proposed system. The \emph{Environmental Engine} is the heart of the system, responsible for the storage and the analysis of the visual content. The content providers are \emph{Mountain Photo Crawler} that retrieves the relevant photographs from all supported public photo repositories, \emph{Mountain Webcam Crawler}, that provides acquired images from all monitored public webcams and the \emph{Front End Web Portal} itself, since users are encouraged to upload their own photographs. \emph{Environmental Engine} constantly processes new content by identifying mountain peaks, estimating the snow coverage and updating it when a manual alignment from a user is provided.
Finally the \emph{Front End Portal} provides the interface to the users and stores all non-environmental information, including social interactions between the users.

Although the technical demonstration consists in the presentation of the \emph{Front End Portal} that allows to appreciate the whole architecture and the interactions between the different components. Apart from features common to any modern photo sharing and social platform (i.e. profile, frienships and followers, comments and likes) the most peculiar actions that the user will be able to perform during the presentation are:

 - Upload an new photograph\ignore{(Supplemental Document, \emph{SD} henceforth, Figure~1),} and inspect the EXIF data useful for the processing. If some data are missing or incorrect (i.e. despite the geo-tag imprecisely saved in the photograph the user is sure about the correct location of the photograph) those can be edited and the photograph is then registered in the system. 
 
 - Explore photographs and webcams on the world map\ignore{ (\emph{SD} Figure~2)}, applying several filters, such as: media type, position, altitude, shot date, mountain peaks contained in. For a deeper exploring of media spatio-temporal frequency distribution, a heat map layer is available\ignore{ (\emph{SD} Figure~3)}.
 
 - Inspect photo details\ignore{ (\emph{SD} Figure~5)}, by interactively exploring the mountain peaks identified on the photo together with the 3D structure of the terrain by providing altitude and distance for each ground point. An environmental mask layer\ignore{ (\emph{SD} Figure~6)} can be turned on showing the classification of the image content in sky/non-relevant terrain (too close to the photographer)/ground/snow areas.
 
 - Inspect webcam details\ignore{ (\emph{SD} Figure~7)}, likewise with the previous page analyzing details of a webcam, and the environmental classified mask\ignore{ (\emph{SD} Figure~8)}. The possibility to start a time-lapse presentation of the webcam images is also implemented. 
 
  - For each photograph or webcam, inspect the automatic alignment performed by the system between the image and the rendered terrain view\ignore{ (\emph{SD} Figure~4)}. If the alignment is wrong ,it can be adjusted through a non-linear warping of the image.

\bibliographystyle{IEEEbib}
\bibliography{icme2015template}

\begin{thebibliography}{1}

\bibitem{garvelmann2013observation}
J~Garvelmann, S~Pohl, and M~Weiler,
\newblock ``From observation to the quantification of snow processes with a
  time-lapse camera network,''
\newblock {\em Hydrology and Earth System Sciences}, vol. 17, pp. 1415--1429,
  2013.

\bibitem{fedorov2014snow}
Roman Fedorov, Piero Fraternali, and Marco Tagliasacchi,
\newblock ``Snow phenomena modeling through online public media,''
\newblock {\em Image Processing, 2014 IEEE International Conference on}, p.
  2179.

\bibitem{fedorov2014mountain}
Roman Fedorov, Piero Fraternali, and Marco Tagliasacchi,
\newblock ``Mountain peak identification in visual content based on coarse
  digital elevation models,''
\newblock in {\em Proceedings of the 3rd ACM International MAED Workshop}.
  2014, MAED '14, ACM.

\end{thebibliography}
\end{document}